\documentclass[runningheads]{llncs}
\usepackage{times}
\usepackage{latexsym}
\usepackage{amsmath}
\usepackage{mathtools}
\usepackage{tikz}
\usepackage{pgf}
\usepackage{natbib}

\usepackage{url}
\usepackage{todonotes}

\usepackage{hyperref}
\usepackage{tcolorbox}

\title{OpenTapioca: Lightweight Entity Linking for Wikidata}
\author{Antonin Delpeuch\inst{1}\orcidID{0000-0002-8612-8827}}

\institute{Department of Computer Science, University of Oxford, UK
\email{antonin.delpeuch@cs.ox.ac.uk}}

\date{}

\newtcbox{\entoure}[1][blue]{on line,
arc=3pt,colback=#1!10!white,colframe=#1!50!black,
before upper={\rule[-3pt]{0pt}{10pt}},boxrule=1pt,
boxsep=0pt,left=2pt,right=2pt,top=1pt,bottom=.5pt}

\definecolor{darkblue}{rgb}{0, 0, 0.5}

\begin{document}

\maketitle

\begin{abstract}
  We propose a simple Named Entity Linking system that can be trained
  from Wikidata only. This demonstrates the strengths and weaknesses
  of this data source for this task and provides an easily
  reproducible baseline to compare other systems against. Our model is
  lightweight to train, to run and to keep synchronous with Wikidata in real
  time.
\end{abstract}

\newcommand{\linkmention}[2]{$\text{\underline{#1}}_{\text{\hspace{0.05em} #2}}$}
\newcommand{\enwikilink}[1]{\href{https://en.wikipedia.org/wiki/#1}{#1}}
\newcommand{\wdlink}[2]{\href{https://www.wikidata.org/entity/#1}{#2 (#1)}}
\newcommand{\qid}[1]{\href{https://www.wikidata.org/entity/#1}{#1}}

\section{Introduction}

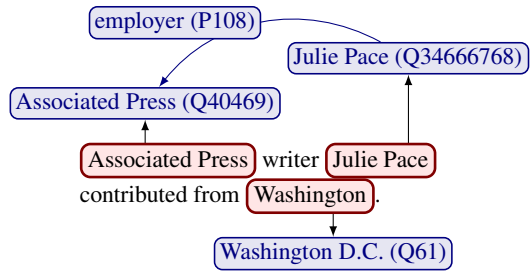
\begin{figure}
  \centering
  \begin{tikzpicture}

    \begin{scope}[every node/.style={rectangle,draw,darkblue,inner sep=2pt,rounded corners=3pt,semithick,fill=darkblue!10}]
    \node at (-1.5,1) (AP) {\wdlink{Q40469}{Associated Press}};
    \node at (2.,1.6) (JP) {\wdlink{Q34666768}{Julie Pace}};
    \node at (1,-1) (DC) {\wdlink{Q61}{Washington D.C.}};
    \draw[-latex,darkblue,bend right=40] (JP) edge node[left] {\wdlink{P108}{employer}} (AP);

    \draw[latex-] (AP) -- (-1.5,0.4);
    \draw[latex-] (JP) -- (2,0.4);
    \draw[latex-] (DC) -- (1,-.4);
    \end{scope}
        \node at (0,0) (text) {\begin{tabular}{l} \entoure[red]{Associated Press} writer \entoure[red]{Julie Pace} \\
        contributed from \entoure[red]{Washington}.
    \end{tabular}};

  \end{tikzpicture}

  \caption{Example of an annotated sentence}
  \label{fig:example-sentence}
\end{figure}

Named Entity Linking is the task of detecting mentions of entities from a
knowledge base in free text, as illustrated in Figure~\ref{fig:example-sentence}.

Most of the entity linking literature focuses on target knowledge
bases which are derived from Wikipedia, such as
DBpedia~\citep{auer2007dbpedia} or YAGO~\cite{suchanek2007yago}. These
bases are curated automatically by harvesting information from the
info-boxes and categories on each Wikipedia page and are therefore not
editable directly.

Wikidata~\cite{vrandevcic2014wikidata} is an editable, multilingual knowledge
base which has recently gained popularity as a target database for
entity linking~\citep{klang2014named,weichselbraun2018mining,sorokin2018mixing,raiman2018deeptype}.
As these new approaches to entity linking also introduce novel
learning methods, it is hard to tell apart the benefits that come
from the new models and those which come from the choice of
knowledge graph and the quality of its data.

We review the main differences between Wikidata and static knowledge
bases extracted from Wikipedia, and analyze their implactions for
entity linking. We illustrate these differences by building a simple
entity linker, OpenTapioca\footnote{The implementation and datasets are available at
  \url{https://github.com/wetneb/opentapioca} and the demo can be
  found at \url{https://opentapioca.org/}.}, which only uses data from
Wikidata, and show that it is competitive with other systems with
access to larger data sources for some tasks. OpenTapioca can be
trained easily from a Wikidata dump only, and can be efficiently kept
up to date in real time as Wikidata evolves. We also propose tools to
adapt existing entity linking datasets to Wikidata, and offer a new
entity linking dataset, consisting of affiliation strings extracted
from research articles.

\section{Particularities of Wikidata} \label{sec:wikidata}

Wikidata is a wiki itself, meaning that it can be edited by anyone,
but differs from usual wikis by its data model: information about an
entity can only be input as structured data, in a format that is
similar to RDF.

Wikidata stores information about the world in a collection of
\emph{items}, which are structured wiki pages. Items are identified by
ther Q-id, such as \qid{Q40469}, and they are made of several data
fields. The \emph{label} stores the preferred name for the entity. It
is supported by a \emph{description}, a short phrase describing the
item to disambiguate it from namesakes, and \emph{aliases} are
alternate names for the entity. These three fields are stored
separately for each language supported by Wikidata. Items also hold a
collection of \emph{statements}: these are RDF-style claims which have
the item as subject. They can be backed by \emph{references} and be
made more precise with \emph{qualifiers}, which all rely on a
controlled vocabulary of \emph{properties} (similar to RDF
predicates).
Finally, items can have \emph{site links}, connecting them to the corresponding page for the
entity in other Wikimedia projects (such as Wikipedia).  Note that Wikidata
items to not need to be associated with any Wikipedia page: in fact, Wikidata's
policy on the notability of the subjects it covers is much more permissive
than in Wikipedia. For a more detailed introduction to Wikidata's data
model we refer the reader to
\citet{vrandevcic2014wikidata,geiss2017neckar}.

Our goal is to evaluate the usefulness of this crowdsourced structured
data for entity linking. We will therefore refrain from augmenting it
with any external data (such as phrases and topical information
extracted from Wikipedia pages), as is generally done when working
with DBpedia or YAGO. By avoiding a complex mash-up of data coming
from disparate sources, our entity linking system is also simpler and
easier to reproduce. Finally, it is possible keep OpenTapioca in
real-time synchronization with the live version of Wikidata, with a
lag of a few seconds only. This means that users are able to fix or
improve the knowledge graph, for instance by adding a missing alias on
an item, and immediately see the benefits on their entity linking
task. This constrasts with all other systems we are aware of, where
the user either cannot directly intervene on the underlying data, or
there is a significant delay in propagating these updates to the
entity linking system.

\section{Related work}

We review the dominant architecture of
entity linking heuristics following \citet{shen2015entity-1},
and assess its applicability to Wikidata.

Entities in the knowledge base are associated with a set (or
probability distribution) of possible surface forms. Given a text to
annotate, candidate entities are generated by looking for occurrences
of their surface forms in the text. Because of homonymy, many of these
candidate occurrences turn out to be false matches, so a classifier is
used to predict their correctness. We can group the features they tend
to use in the following categories:
\begin{itemize}
\item \textbf{local compatibility}: these features assess the adequacy
  between an entity and the phrase that refers to it. This relies
  on the dictionary of surface forms mentioned above, and does not
  take into account the broader context of the phrase to link.
\item \textbf{topic similarity}: this measures the compatibility
  between the topics in the text to annotate and the topics
  associated with the candidate entity.  Topics can be represented in
  various ways, for instance with a bag of words model.
\item \textbf{mapping coherence}: entities mentioned in the same text
  are often related, so linking decisions are inter-dependent. This relies
  on a notion of proximity between entities, which can be defined with
  random walks in the knowledge graph for instance.
\end{itemize}

 \subsection{Local compatibility}

 These features compare the phrase to annotate with the known surface forms
 for the entity. Collecting such forms is often done by extracting
 mentions from Wikipedia~\citep{cucerzan2007large}. Link labels,
 redirects, disambiguation pages and bold text in abstracts can all
 be useful to discover alternate names for an entity. It is also possible
 to crawl the web for Wikipedia links to improve the coverage,
 often at the expense of data quality~\citep{spitkovsky2012crosslingual}.

 Beyond collecting a set of possible surface forms, these approaches
 count the number of times an entity $e$ was mentioned by a phrase
 $w$.  This makes it possible to use a Bayesian methodology: the
 compatibility of a candidate entity $e$ with a given mention $w$ is
 $P(e | w) = \frac{P(e,w)}{P(w)}$, which can be estimated from the
 statistics collected.
 
 In Wikidata, items have labels and aliases in multiple languages.  As
 this information is directly curated by editors, these phrases tend
 to be of high quality. However, they do not come with occurence
 counts. As items link to each other using their Wikidata identifiers
 only, it is not possible to compare the number of times \texttt{USA}
 was used to refer \wdlink{Q30}{United States of America} or to
 \wdlink{Q9212}{United States Army} inside Wikidata.

 Unlike Wikipedia's page titles which must be
 unique in a given language, two Wikidata items can have the same label in the same
 language. For instance \texttt{Curry} is the English label of both
 the item about the Curry programming language (\qid{Q2368856}) and
 the item about the village in Alaska (\qid{Q5195194}), and the
 description field is used to disambiguate them.

 Manual curation of surface forms implies a fairly narrow coverage,
 which can be an issue for general purpose entity linking. For
 instance, people are commonly refered to with their given or family
 name only, and these names are not systematically added as aliases: at
 the time of writing, \texttt{Trump} is an alias for
 \wdlink{Q22686}{Donald Trump}, but \texttt{Cameron} is not an alias for
 \wdlink{Q192}{David Cameron}. As a Wikidata editor, the main
 incentive to add aliases to an item is to make it easier to find the item with
 Wikidata's auto-suggest field, so that it can be edited or linked to
 more easily. Aliases are not designed to offer a complete set of
 possible surface forms found in text: for instance, adding common mispellings
 of a name is discouraged.\footnote{The guidelines are available at \url{https://www.wikidata.org/wiki/Help:Aliases}}
 
 \subsection{Topic similarity}

 The compatibility of the topic of a candidate entity with the rest
 of the document is traditionally estimated by similarity measures
 from information retrieval such as
 TFIDF~\citep{vstajner2009entity,ratinov2011local} or keyword
 extraction~\citep{strube2006wikirelate,mihalcea2007wikify,cucerzan2007large}.
 
 Wikidata items only consist of structured data, except in their
 descriptions. This makes it difficult to compute topical information
 using the methods above. Vector-based representations of entities can
 be extracted from the knowledge graph
 alone~\citep{bordes2013translating,xiao2016transg}, but it is not
 clear how to compare them to topic representations for plain text,
 which would be computed differently.  In more recent work, neural
 word embeddings were used to represent topical information for both
 text and
 entities~\citep{ganea2017deep,raiman2018deeptype,kolitsas2018endtoend}. This
 requires access to large amounts of text both to train the word
 vectors and to derive the entity vectors from them. These vectors
 have been shown to encode significant semantic information by
 themselves~\citep{mikolov2013efficient}, so we refrain from using
 them in this study.

\subsection{Mapping coherence}

Entities mentioned in the same context are often topically related,
therefore it is useful not to treat linking decisions in isolation
but rather to try to maximize topical coherence in the chosen
items. This is the issue on which entity linking systems differ
the most as it is harder to model.

First, we need to estimate the topical coherence of a sequence of
linking decisions. This is often done by first defining a pairwise
relatedness score between the target entities. For instance, a popular
metric introduced by \citet{witten2008effective} considers the set of
wiki links $|a|, |b|$ made from or to two entities $a$, $b$ and
computes their relatedness:
$$ \text{rel}(a,b) = 1 - \frac{\log(\max(|a|,|b|)) - \log(|a| \cap |b|)}{\log(|K|) - \log(\min(|a|,|b|))}$$
where $|K|$ is the number of entities in the knowledge base.

When linking to Wikidata instead of Wikipedia, it is tempting to reuse
these heuristics, replacing wikilinks by statements.  However,
Wikidata's linking structure is quite different from Wikipedia:
statements are generally a lot sparser than links and they have a
precise semantic meaning, as editors are restricted by the available
properties when creating new statements. We propose in the next
section a similarity measure that we find to perform well
experimentally.

Once a notion of semantic similarity is chosen, we need to integrate
it in the inference process.
Most approaches build a graph of candidate entities, where
edges indicate semantic relatedness: the difference between the
heuristics lie in the way this graph is used for the matching
decisions.  \citet{moro2014entity} use an approximate algorithm to
find the \emph{densest subgraph} of the semantic graph. This
determines choices of entities for each mention.  In other approaches,
the initial evidence given by the local compatibility score is
propagated along the edges of the semantic graph
\citep{mihalcea2007wikify,han2011collective} or aggregated
at a global level with a Conditional Random Field~\citep{ganea2017deep}.

\section{OpenTapioca: an entity linking model for Wikidata}

We propose a model that adapts previous approaches to Wikidata.  Let
$d$ be a document (a piece of text). A \emph{spot} $s \in d$ is a pair
of start and end positions in $d$. It defines a phrase $d[s]$, and a
set of candidate entities $E[s]$: those are all Wikidata items for
which $d[s]$ is a label or alias. Given two spots $s, s'$ we denote by
$|s - s'|$ the number of characters between them. We build a binary
classifier which predicts for each $s \in d$ and $e \in E[s]$ if $s$
should be linked to $e$.

\subsection{Local compatibility}

Although Wikidata makes it impossible to count how often a particular
label or alias is used to refer to an entity, these surface forms are
carefully curated by the community. They are therefore fairly reliable.

Given an entity $e$ and a phrase $d[s]$, we need to compute $p(e|
d[s])$.  Having no access to such a probability distribution, we
choose to approximate this quantity by $\frac{p(e)}{p(d[s])}$, where
$p(e)$ is the probability that $e$ is linked to, and $p(d[s])$ is the
probability that $d[s]$ occurs in a text. In other words, we estimate
the popularity of the entity and the commonness of the phrase separately.

We estimate the popularity of an entity $e$ by a log-linear
combination of its number of statements $n_e$, site links $s_e$ and
its PageRank $r(e)$. The PageRank is computed on the entire Wikidata
using statement values and qualifiers as edges.

The probability $p(d[s])$ is estimated by a simple unigram language model
that can be trained either on any large
unannotated dataset\footnote{For the sake of respecting our constraint
  to use Wikidata only, we train this language model from Wikidata
  item labels.}.

The local compatibility is therefore represented by a vector of features
$F(e,w)$ and the local compatibility is computed as follows, where $\lambda$
is a weights vector:
\begin{align*}
  F(e,w) &= ( -\log p(d[s]), \log p(e) , n_e, s_e, 1 ) \\
  p(e|d[s]) &\propto e^{F(e,w) \cdot \lambda}
\end{align*}

\subsection{Semantic similarity}

The issue with the features above is that they ignore the context in
which a mention in found. To make it context-sensitive, we adapt the
approach of \citet{han2011collective} to our setup. The general idea
is to define a graph on the candidate entities, linking candidate
entities which are semantically related, and then find a combination
of candidate entities which have both high local compatibility and
which are densely related in the graph.

For each pair of entities $e, e'$ we define a similarity metric $s(e,e')$. Let $l(e)$ be the set of items that $e$ links to in its statements. Consider a one-step random walks starting on $e$, with probability $\beta$ to stay on $e$ and probability $\frac{1-\beta}{|l(e)|}$ to reach one of the linked items. We define $s(e,e')$ as the probability that two such one-step random walks starting from $e$ and $e'$ end up on the same item. This can be computed explicitly as
\begin{align*}s(e,e') &= \beta^2 \delta_{e=e'} + \beta(1 - \beta)(\frac{\delta_{e \in l(e')}}{|l(e')|} \\
  &+ \frac{\delta_{e' \in l(e)}}{|l(e)|}) + (1 - \beta)^2 \frac{|l(e) \cap l(e')|}{|l(e)||l(e')|}
\end{align*}
In this formula, $\delta_P$ stands for $1$ when $P$ is true and $0$ otherwise.

We then build a weighted graph $G_d$ whose vertices are pairs $(s \in
d, e \in E[s])$. In other words, we add a vertex for each candidate entity at a given spot.
We fix a maximum distance $D$ for edges: vertices $(s,e)$ and $(s',e')$ can only
be linked if $|s - s'| \leq D$ and $s \neq s'$.
In this case, we define the weight of such an edge as $(\eta + s(e,e'))\frac{D - |s - s'|}{D}$,
where $\eta$ is a smoothing parameter. In other words, the edge weight is proportional
to the smoothed similarity between the entities, discounted by the distance between the mentions.

The weighted graph $G_d$ can be
represented as an adjacency matrix. We transform it into a
column-stochastic matrix $M_d$ by normalizing its columns to sum to one.
This defines a Markov chain on the candidate entities, that we will
use to propagate the local evidence.

\subsection{Classifying entities in context}

\citet{han2011collective} first combine the local features into a
local evidence score, and then spread this local evidence using the
Markov chain:
\begin{align} \label{eqn:g-def}
G(d) = (\alpha I + (1 - \alpha) M_d)^k \cdot LC(d)
\end{align}
We propose a variant of this approach, where each individual local
compatibility feature is propagated independently along the Markov
chain.\footnote{It is important for this purpose that features are
  initially scaled to the unit interval.}  Let $F$ be the matrix of
all local features for each candidate entity: $F = (F(e_1,d[s_1]),
\dots, F(e_n, d[s_n]))$.  After $k$ iterations in the Markov chain,
this defines features $M_d^k F$. Rather than relying on these
features for a fixed number of steps $k$, we record the features at
each step, which defines the vector
$$(F, M_d \cdot F, M_d^2 \cdot F, \dots, M_d^k \cdot F)$$ This
alleviates the need for an $\alpha$ parameter while keeping the number
of features small. We train a linear support vector classifier on
these features and this defines the final score of each candidate
entity. For each spot, our system picks the highest-scoring candidate
entity that the classifier predicts as a match, if any.

\section{Experimental setup}

Most entity linking datasets are annotated against DBpedia or YAGO.
Wikidata contains items which do not have any corresponding Wikipedia
article (in any language), so these items do not have any DBpedia or
YAGO URI either.\footnote{This is the case of \wdlink{Q34666768}{Julie Pace} in Figure~\ref{fig:example-sentence}.} Therefore, converting an entity linking dataset from DBpedia
to Wikidata requires more effort than simply following
\texttt{owl:sameAs} links: we also need to annotate mentions of
Wikidata items which do not have a corresponding DBpedia URI.

We used the RSS-500 dataset of news excerpts annotated against
DBpedia and encoded in NIF format~\citep{usbeck2015gerbil}.  We first translated all
DBpedia URIs to Wikidata items\footnote{We built the
  \href{https://github.com/wetneb/nifconverter}{nifconverter} tool to do this conversion for any NIF
  dataset.}. Then, we
used OpenRefine~\citep{openrefine} to extract the entities marked not covered by
DBpedia and matched them against Wikidata. After human review, this
added 63 new links to the 524 converted from DBpedia (out of 476
out-of-KB entities).

We also annotated a new dataset from scratch. The ISTEX dataset
consists of one thousand author affiliation strings extracted from
research articles and exposed by the ISTEX text and data mining
service\footnote{The original data is available under an Etalab
  license at \url{https://www.istex.fr/}}. In this dataset, only 64 of the
2,624 Wikidata mentions do not have a corresponding DBpedia URI.

We use the Wikidata JSON dump of 2018-02-24 for our experiments,
indexed with Solr (Lucene). We restrict the index to
humans, organizations and locations, by selecting only items whose
type was a \wdlink{P279}{subclass of} \wdlink{Q5}{human},
\wdlink{Q43229}{organization} or \wdlink{Q618123}{geographical
  object}. Labels and aliases in all languages are added to a
case-sensitive FST index.

We trained our classifier and its hyper-parameters by five-fold
cross-validation on the training sets of the ISTEX and RSS datasets.
We used GERBIL~\citep{usbeck2015gerbil} to evaluate OpenTapioca
against other approaches.  We report the InKB micro and macro F1
scores on test sets, with GERBIL's weak annotation match
method.\footnote{The full details can be found at
  \url{http://w3id.org/gerbil/experiment?id=201904110006}}
\begin{figure}
\centering
{\small
\hspace{-.5cm}
\begin{tabular}{l c c c c c c c c}
  & \multicolumn{2}{c}{AIDA-CoNLL} & \multicolumn{2}{c}{Microposts 2016} \\
  & Micro & Macro & Micro & Macro \\
  \hline
  AIDA &          \textbf{0.725} & \textbf{0.684} & 0.056 & 0.729 \\
  Babelfy &       0.481 & 0.421 & 0.031 & 0.526 \\
  DBP Spotlight & 0.528 & 0.456 & 0.053 & 0.306 \\
  FREME NER &     0.382 & 0.237 & 0.037 & \textbf{0.790} \\
  OpenTapioca &   0.482 & 0.399 & \textbf{0.087} & 0.515\\
  \hline
  & \multicolumn{2}{c}{ISTEX-1000} & \multicolumn{2}{c}{RSS-500} \\
  & Micro & Macro & Micro & Macro \\
  \hline
  AIDA &          0.531 & 0.494 & \textbf{0.455} & \textbf{0.447} \\
  Babelfy &       0.461 & 0.447 & 0.314 & 0.304 \\
  DBP Spotlight & 0.574 & 0.575 & 0.281 & 0.261 \\
  FREME NER &     0.422 & 0.321 & 0.307 & 0.274 \\
  OpenTapioca &   \textbf{0.870} & \textbf{0.858} & 0.335 & 0.310 \\
  \hline
\end{tabular}
}
\caption{F1 scores on test datasets}
\end{figure}

\section{Conclusion}

The surface forms curated by Wikidata editors are sufficient to reach
honourable recall, without the need to expand them with mentions
extracted from Wikipedia.  Our restriction to people, locations and
organizations probably helps in this regard and we anticipate worse
performance for broader domains. Our approach works best for
scientific affiliations, where spelling is more canonical than in
newswire.  The availability of Twitter identifiers directly in
Wikidata helps us to reach acceptable performance in this domain. The
accuracy degrades on longer texts which require relying more on the
ambiant topical context. In future work, we would like to explore the
use of entity embeddings to improve our approach in this regard.

\bibliographystyle{acl_natbib}
\bibliography{zotero}

\end{document}